\documentclass[lettersize,journal]{IEEEtran}
\usepackage{amsmath,amsfonts}
\usepackage{mathtools}
\usepackage{algorithmic}
\usepackage{algorithm}
\usepackage{array}
\usepackage[caption=false,font=normalsize,labelfont=sf,textfont=sf]{subfig}
\usepackage{textcomp}
\usepackage{stfloats}
\usepackage{url}
\usepackage{verbatim}
\usepackage{graphicx}
\usepackage{cite}
\usepackage{accents}
\usepackage{balance}
\usepackage[colorlinks=true,linkcolor=black,citecolor=black,urlcolor=blue]{hyperref}

\makeatletter
\newcommand{\raisemath}[1]{\mathpalette{\raiseMath{#1}}}%
\newcommand{\raiseMath}[3]{\raisebox{#1}[0pt][0pt]{$#2#3$}}
\makeatother

\newcommand{\largedot}[2][1.4]{
\accentset{\scalebox{#1}{$\mbox{.}$}}{#2}
}

\NewDocumentCommand{\qbar}{O{0.5pt} O{-6.55pt}}{
	\ensuremath{\mathrlap{\raisemath{#2}{\hspace*{#1}{\mathchar'26\mkern-9mu}}} q}%
}

\NewDocumentCommand{\qbars}{O{0.5pt} O{-4.65pt}}{
	\ensuremath{\mathrlap{\raisemath{#2}{\hspace*{#1}{\mathchar'26\mkern-9mu}}} q}%
}

\NewDocumentCommand{\qbarc}{O{0.5pt} O{-5.2pt}}{
	\ensuremath{\mathrlap{\raisemath{#2}{\hspace*{#1}{\mathchar'26\mkern-9mu}}} q}%
}

\NewDocumentCommand{\pbar}{O{-1.5pt} O{-6.65pt}}{
	\ensuremath{\mathrlap{\raisemath{#2}{\hspace*{#1}{\mathchar'26\mkern-9mu}}} p}%
}

\newcommand{\bs}[1]{\boldsymbol{#1}}  
\newcommand{\ts}[1]{\text{#1}} 

\renewcommand{\baselinestretch}{0.98}  

\begin{document}
\author{}

\title{\fontsize{24pt}{23pt}\selectfont Model-Reference Adaptive Flight Control of a \mbox{95-mg} \mbox{Insect-Scale} \mbox{Flapping-Wing} Aerial Robot}

\author{Francisco M. F. R. Gon\c{c}alves, Conor K. Trygstad, and N\'{e}stor O. P\'{e}rez-Arancibia
\thanks{This work was supported in part by the Washington State University (WSU) Foundation and the Palouse Club through a Cougar Cage Award to N.\,O.\,P\'erez-Arancibia and in part by the WSU Voiland College of Engineering and Architecture through a start-up fund to N.\,O.\,P\'erez-Arancibia.}
\thanks{The authors are with the School of Mechanical and Materials Engineering, Washington State University (WSU), Pullman, WA 99164-2920, USA. Corresponding authors' email: {\tt francisco.goncalves@wsu.edu} (F.\,M.\,F.\,R.\,G.); {\tt n.perezarancibia@wsu.edu} (N.\,O.\,P.-A.).}}



\maketitle

\begin{abstract}
 Due to the system's scale and complex fabrication, the model describing the dynamics of a \mbox{flapping-wing} \mbox{insect-scale} aerial robot is subject to parameter uncertainty; for example, in the inertia matrix and the actuator mapping of the flier. Furthermore, due to its low inertia, this type of robot is greatly affected by stochastic and systematic disturbances during flight, including \mbox{power-wire} tension, gusts, and undesired aerodynamic forces produced by wing misalignment. Therefore, the \mbox{high-performance} execution of complex maneuvers at the sub-decigram scale requires the robot to adapt its behavior to counteract disturbances and model uncertainty. Toward this objective, we introduce a \textit{model-reference adaptive control} (MRAC) architecture for \mbox{high-performance} position control of \mbox{flapping-wing} robotic insects that can be modeled as rigid bodies in the \mbox{\textit{three-dimensional}}~($\bs{3}$D) space. In addition, we demonstrate how the implementation of a hybrid multiplicative extended K\'{a}lm\'{a}n filter for estimating current and desired angular velocities during flight significantly dampens attitude vibrations, especially along the roll and pitch \textit{degrees of freedom} (DOFs), and also improves flight performance. To show the suitability, functionality, and high performance of the proposed approach, we conducted \mbox{real-time} hovering and \mbox{trajectory-tracking} $\bs{6}$-DOF flight control experiments with a \mbox{$\bs{95}$-mg} \mbox{insect-scale} aerial robot.
\end{abstract}

\begin{IEEEkeywords}
Bioinspired robots, flight control, microrobotics.
\end{IEEEkeywords}

\section{Introduction}
\begin{figure}
    \vspace{0.4ex}
    \begin{center}
    \includegraphics[width=3.4in]{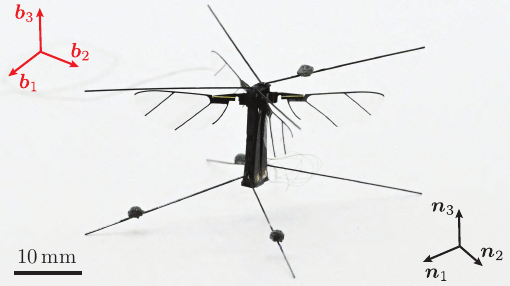}
    \end{center}
    \caption{\textbf{Photograph of the \mbox{$\bs{95}$-mg} \mbox{insect-scale} aerial robot used in the flight experiments.} The robot and the definitions of the \mbox{body-fixed} and inertial frames, \mbox{$\bs{\mathcal{B}} = \left\{\bs{b}_1,\bs{b}_2,\bs{b}_3\right\}$} and \mbox{$\bs{\mathcal{N}} = \left\{\bs{n}_1,\bs{n}_2,\bs{n}_3\right\}$}, respectively. \label{Fig01}} 
    \vspace{-1ex}
\end{figure}

\IEEEPARstart{R}{ecent} advances in \mbox{insect-scale} \mbox{flapping-wing} robots have enabled effective flight control of the six \textit{degrees of freedom}~(DOFs)~\cite{BenaRM2023I}. The ability to control the yaw DOF during flight has opened the doors for new types of tasks for these systems, such as visual navigation, object tracking, and collision avoidance systems~\cite{BenaRM2023II}, where yaw tracking is essential to orient the robot toward a desired direction of interest. More recently, researchers have developed a \mbox{$750$-mg} robot of the same type and demonstrated its high agility and ability to perform \mbox{high-speed} aerobatic maneuvers\mbox{\cite{HsiaoYH2025, KimS2025, HsiaoYH2025II}}. Despite its impressive flight performance and conceivable advantages, this robot is almost $8$ times heavier and $5$ times larger (in volume) than the robot used in the experiments reported in this paper, which limits its applicability to situations where being unnoticed and undetectable is imperative, and makes it unsuitable for navigating spaces narrower than $4$\,cm due to its dimensions. Furthermore, the robot in\mbox{\cite{HsiaoYH2025, KimS2025, HsiaoYH2025II}} cannot generate yaw torque, thereby limiting its capabilities in many \mbox{real-world} applications. Therefore, the \mbox{$95$-mg} robot used in the experiments reported in this paper is still the only \mbox{insect-scale} flying robot capable of effectively controlling its six DOFs while achieving \mbox{high-performance} flight. This microrobot---shown in \mbox{Fig.\,\ref{Fig01}}---is driven by four unimorph piezoelectric actuators powered via a tether wire~\cite{yang2019bee}. Due to the system's scale and complex fabrication, the model describing the associated dynamics is subject to parameter uncertainty, mainly in the inertia matrix and the actuator mapping of the flier, which directly affect the directions of the generated torques for attitude and position control. Furthermore, due to its low inertia, this robot is greatly affected by stochastic and systematic disturbances during flight, including \mbox{power-wire} tension, gusts, and undesired aerodynamic forces produced by wing misalignment during fabrication. Because the disturbance forces are often time varying and unknown to the controller designer, neither \mbox{integral-type} nor \mbox{gain-scheduling} control architectures are well suited to compensate for them during flight~\cite{Ioannou_AdaptiveControl}. Thus, an adaptive control structure is a more appropriate choice for this scenario.

While the adaptive control literature has a vast number of examples showing applications to \mbox{large-scale} \textit{uncrewed aerial vehicles}~(UAVs) (see~\cite{WangJ2023, WhiteheadBT2010,WiseKA2006,DydekZT2013,YinY2024,MoH2018} and references therein), the amount of published research with applications to \mbox{insect-scale} \mbox{flapping-wing} aerial robots is scarce\mbox{\cite{ArancibiaNP2013, chirarattananon2013adaptive, chirarattananon2014adaptive,JimboT2026}}. Specifically, in~\cite{ArancibiaNP2013}, the authors introduce an adaptive feedforward cancelation algorithm for the problem of lift-force control for \mbox{insect-scale} aerial robots. However, at the time, position control in the \mbox{\textit{three-dimensional}}~($3$D) space with this type of robot had not yet been achieved, and, therefore, the approach is limited to altitude control with the robot constrained to move only in the vertical direction and to follow simple trajectories. In~\cite{chirarattananon2013adaptive} and~\cite{chirarattananon2014adaptive}, the authors present an adaptive control approach to address model uncertainties. Despite its impressive performance---using a different platform from the one used in the work presented in this paper---measured in terms of the \mbox{\textit{root-mean-square}}~(RMS) position error and compared to a nonadaptive controller implemented on the same platform, this robot is still not capable of controlling its yaw DOF. As demonstrated by flying insects in nature~\cite{BoeddekerN2010, BoeddekerN2010II}, this is a fundamental capability these robots must possess to be able to perform \mbox{real-world} complex tasks in the future. More recently~\cite{JimboT2026}, researchers have developed a new \mbox{insect-scale} \mbox{flapping-wing} aerial robot capable of yaw control and implemented an adaptive controller to account for lift force offset. However, only one controlled flight experiment with a duration of less than $2$\,seconds is reported, which shows that this robot is not yet capable of sustained flight.

In this paper, we present the first implementation of a \textit{\mbox{model-reference} adaptive control}~(MRAC) architecture on a \mbox{flapping-wing} robotic insect for \mbox{high-performance} position control. In this approach, we combine the multiplatform position control scheme proposed in~\cite{BenaRM2023I}~and~\cite{BenaRM2022} with an MRAC architecture for estimating and canceling the unknown aggregated disturbance affecting the robot. Moreover, because the control of \mbox{insect-scale} aerial robots still relies on \mbox{motion-capture} systems for sensing, unlike most UAVs---which get direct \mbox{angular-velocity} measurements using an \textit{inertial measurement unit} (IMU)---these microrobots get direct measurements of their orientation while their angular rate needs to be estimated. Previously, in~\cite{BenaRM2023I}, researchers have used a \mbox{low-pass} derivative filter for this purpose; however, this results in extremely noisy estimates and, as a consequence, \mbox{relatively-large} oscillations during flight. In this work, we also show that using a hybrid multiplicative extended K\'{a}lm\'{a}n filter for estimating both the desired and current angular velocities of the robot significantly dampens attitude vibrations and improves overall performance measured in terms of RMS position-tracking errors. The \mbox{high-performance} of the proposed approach is validated through multiple \mbox{real-time} $6$-DOF hovering and \mbox{trajectory-tracking} control experiments using a \mbox{$95$-mg} \mbox{insect-scale} \mbox{flapping-wing} aerial robot.

The rest of the paper is organized as follows. Section\,\ref{Section02} discusses the modeling of the translational dynamics of the robot and aggregated disturbance. Section\,\ref{Section03} presents the design of the adaptive law derived from Lyapunov stability theory. Section\,\ref{Section04} discusses the new method used for the estimation of the current and desired angular velocities of the robot during flight; presents hovering and \mbox{trajectory-tracking} flight experiments using a \mbox{$95$-mg} \mbox{insect-scale} aerial robot; and compares its performance with the nonadaptive version of the same controller as well as with previous published results using the same platform. Last, Section\,\ref{Section05} draws some conclusions from the presented results and discusses possible paths for future research.

\vspace{1ex}
\textit{\textbf{Notation--}}
\begin{enumerate}
\item Italic lowercase symbols represent scalars, e.g., $p$; bold lowercase symbols represent vectors, e.g., $\bs{p}$; bold uppercase symbols represent matrices, e.g., $\bs{P}$; and bold crossed lowercase symbols represent quaternions, \mbox{e.g., $\bs{\pbar}$}.
\item The real variable $t$ denotes continuous time. The integer variable $k$ is used to index discrete time.
\item The $2$-norm of a vector is denoted by $\| \cdot \|_2$.
\item The $\mathcal{L}_2$-norm of a signal is denoted by~\mbox{$\| \cdot \|_{\mathcal{L}_2}$}.
\item The $\mathcal{L}_{\infty}$-norm of a signal is denoted by~\mbox{$\| \cdot \|_{\mathcal{L}_{\infty}}$}.
\item The bold symbol $\bs{1}$ represents a vector with every entry corresponding to $1$.
\end{enumerate}

\section{Rigid-Body Dynamics \& Disturbance Modeling}\label{Section02}
Using Newton's second law and assuming a rigid body, the translational dynamics of the robot moving in the $3$D space can be described as
\begin{align}
    m\bs{\ddot{r}}(t) = \bs{f}_{\hspace{-0.1ex}\ts{t}}(t) - mg\bs{n}_{3} + \bs{d}(t),
    \label{EQ5.01}
\end{align}
where $m$ is the mass of the robot; \mbox{$\bs{r} = [r_1~r_2~r_3]^T$} is the position of the robot's \textit{center of mass} (CoM) relative to \mbox{$\bs{\mathcal{N}} = \left\{\bs{n}_1,\bs{n}_2,\bs{n}_3\right\}$}, as defined in \mbox{Fig.\,\ref{Fig01}}; $\bs{f}_{\hspace{-0.1ex}\ts{t}}$ is the total force generated by the flapping wings of the robot; $g$ is the acceleration of gravity; and $\bs{d}$ is the unknown aggregated disturbance affecting the flier. We assume that this disturbance can be modeled as
\begin{align}
    \bs{d}(t) = \bs{W}^T\bs{\phi}(\bs{x}),
    \label{EQ5.02}
\end{align}
where $\bs{W}$ is a constant \mbox{$n \times 3$} matrix of unknown parameters and $\bs{\phi}(\bs{x})$ is a vector of \mbox{Gaussian-based} \textit{radial basis functions} (RBFs), in which the $i$th element is given by
\begin{align}
    \phi_i = \ts{exp}\left(\frac{\|\bs{x} - \bs{c}_i\|^2_2}{2\sigma_i^2}\right),~~ i \in \left\{1,2,\cdots,n \right\},
    \label{EQ5.03}
\end{align}
with \mbox{$\bs{x} = [\bs{r}^T~\bs{\dot{r}}^T]^T$}. The real vector $\bs{c}_i$ and the real number $\sigma_i$ define the center and bandwidth of $\phi_i$, respectively.
\begin{figure*}[t!]
    \centering
    \includegraphics[width=\textwidth]{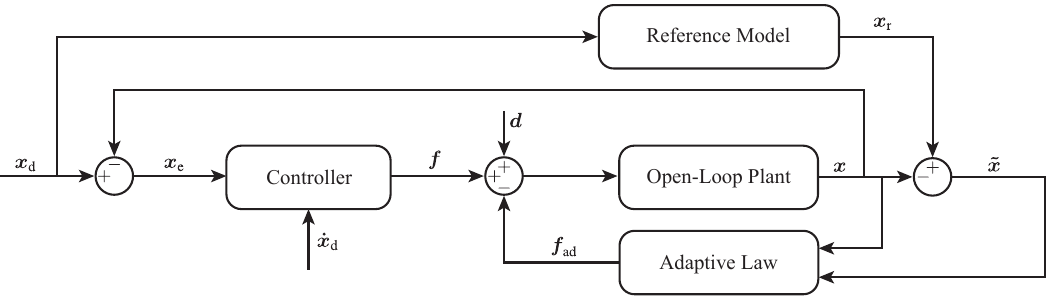}
    \caption{\textbf{Block diagram of the model-reference adaptive control scheme for the translational dynamics.} In this scheme, the reference model generates the desired response of the system, $\bs{x}_{\ts{r}}$, to a given desired trajectory $\bs{x}_{\ts{d}}$. Assuming full measurement state, the adaptive law takes the measured state and the error between the reference model and measured states, $\bs{\Tilde{x}}$, to estimate the adaptive force that aims to cancel the unknown disturbance, $\bs{f}_{\hspace{-0.2ex}\ts{ad}}$. The controller receives the state error between the reference trajectory and the measured state, and computes the nominal force, $\bs{f}$.}
    \label{Fig02}
    \vspace{-3ex}
\end{figure*}

\section{Adaptation and Stability Analysis}\label{Section03}
By design, the robot can only generate a force with magnitude $f_{\hspace{-0.1ex}\ts{t}}(t)$ in the $\bs{b}_3$ direction, as defined in \mbox{Fig.\,\ref{Fig01}}; however, similarly to the case presented in~\cite{BenaRM2023I}, for the purpose of stability analysis, we assume that the attitude controller of the system is fast enough such that \mbox{$f_{\hspace{-0.1ex}\ts{t}}\bs{b}_3 \approx \bs{f}_{\hspace{-0.1ex}\ts{t}}$}. Detailed design and stability analysis of the attitude controller used in this work can be found in~\cite{BenaRM2023I}; however, this is not in the scope of this paper. For a given desired state $\bs{x}_{\ts{d}}$, we define the tracking error to be minimized as \mbox{$\bs{x}_{\ts{e}} = \bs{x}_{\ts{d}} - \bs{x}$} and the corresponding stabilizing control law as
\begin{align}
    \bs{f}_{\hspace{-0.1ex}\ts{t}} = \bs{K}_{\ts{p}} \bs{r}_{\ts{e}} + \bs{K}_{\ts{i}}\int_0^t \bs{r}_{\ts{e}}(\zeta)d\zeta  + \bs{K}_{\ts{d}} \bs{\largedot{r}}_{\ts{e}} + m\bs{\ddot{r}}_{\ts{d}} + mg\bs{n}_3 - \bs{f}_{\hspace{-0.2ex}\ts{ad}},
    \label{EQ5.04}
\end{align}
where $\bs{K}_{\ts{lqr}} = \left[\bs{K}_{\ts{i}}~\bs{K}_{\ts{p}}~\bs{K}_{\ts{d}}\right]$ can be determined using a \textit{linear quadratic regulator} (LQR) design approach;  \mbox{$\bs{r}_{\ts{e}} = \bs{r}_{\ts{d}} - \bs{r}$} is the position error, with $\bs{r}_{\ts{d}}$ smooth and bounded; and, \mbox{$\bs{f}_{\hspace{-0.2ex}\ts{ad}} = \bs{\hat{W}}^T\bs{\phi}(\bs{x})$} is an adaptive term included with the objective of canceling the unknown disturbance affecting the system. In this approach, an adaptive law updates $\bs{\hat{W}}$ such that the linear combination of Gaussian kernels matches the unknown disturbance to be canceled. By substituting~\eqref{EQ5.04} into~\eqref{EQ5.01}, we get
\begin{align}
   \bs{\ddot{r}}_{\ts{e}} = -\frac{1}{m}\left[\bs{K}_{\ts{p}} \bs{r}_{\ts{e}} + \bs{K}_{\ts{i}}\int_0^t \bs{r}_{\ts{e}}(\zeta)d\zeta  + \bs{K}_{\ts{d}} \bs{\dot{r}}_{\ts{e}}\right] -\frac{1}{m} \left(\bs{d} - \bs{f}_{\hspace{-0.2ex}\ts{ad}}\right),
    \label{EQ5.05}
\end{align}
which, by defining \mbox{$\bs{\xi} = \int_0^t \bs{r}_{\ts{e}}(\zeta)d\zeta$}, can be written as 
\begin{align}
    \bs{\dddot{\xi}} = -\frac{1}{m}\left[\bs{K}_{\ts{p}} \bs{\dot{\xi}} + \bs{K}_{\ts{i}}\bs{\xi}  + \bs{K}_{\ts{d}} \bs{\ddot{\xi}}\right] -\frac{1}{m} \left(\bs{d} - \bs{f}_{\hspace{-0.2ex}\ts{ad}}\right).
    \label{EQ5.06}
\end{align}
Furthermore, \eqref{EQ5.06} can be put into the \mbox{state-space} form
\begin{align}
\begin{split}
\bs{\dot{z}}(t) &= \bs{A} \bs{z}(t) + \bs{B}\left(\bs{d} - \bs{f}_{\hspace{-0.2ex}\ts{ad}}\right),
\end{split}
\label{EQ5.07}
\end{align}
where
\begin{align*}
\begin{split}
\bs{z} &= \begin{bmatrix}
\bs{\xi}^T & \bs{\dot{\xi}}^T & \bs{\ddot{\xi}}^T
\end{bmatrix}^T; \\[1ex]
\bs{A} &= \begin{bmatrix}
\bs{0}_{3\times 3} & \bs{I}_{3\times 3} & \bs{0}_{3\times 3}\\[1ex]
\bs{0}_{3\times 3} & \bs{0}_{3\times 3} & \bs{I}_{3\times 3}\\[1ex]
-\frac{1}{m}\bs{K}_{\ts{i}} & -\frac{1}{m}\bs{K}_{\ts{p}} & -\frac{1}{m}\bs{K}_{\ts{d}}
\end{bmatrix}; \\[1ex]
\bs{B} &= \begin{bmatrix}
\bs{0}_{6\times 3}\\[1ex]
-\frac{1}{m}\bs{I}_{3\times 3}
\end{bmatrix}.
\end{split}
\end{align*}
Then, considering \eqref{EQ5.07}, we define a \mbox{zero-disturbance} reference model of the closed-loop system with the form
\begin{align}
\bs{\dot{z}}_{\ts{r}}(t) &= \bs{A} \bs{z}_{\ts{r}}(t).
\label{EQ5.08}
\end{align}
Thus, the \mbox{modeling-error} dynamics are given by
\begin{align}
\begin{split}
  \largedot{\Tilde{\bs{z}}} &= \bs{A}\Tilde{\bs{z}} + \bs{B}\left(\bs{f}_{\hspace{-0.2ex}\ts{ad}} - \bs{d}\right)\\
   &= \bs{A}\Tilde{\bs{z}} + \bs{B}\left(\bs{\hat{W}}^T\bs{\phi}(\bs{x}) - \bs{W}^T\bs{\phi}(\bs{x})\right)\\
   &= \bs{A}\Tilde{\bs{z}} + \bs{B}\bs{\Tilde{W}}^T\bs{\phi}(\bs{x}).
   \end{split}
   \label{EQ5.09}
\end{align} 

From \eqref{EQ5.09}, it follows that an effective adaptive method for rejecting the disturbances affecting the system must produce a matrix $\bs{\hat{W}}$ such that $\lim_{t \rightarrow \infty} \bs{z}(t) = \bs{z}_{\ts{r}}(t)$. To this end, we first define the Lyapunov function candidate
\begin{align}
    V = \Tilde{\bs{z}}^T \bs{P}\Tilde{\bs{z}} + \gamma^{-1}\ts{tr}\left\{\bs{\Tilde{W}}^T\bs{\Tilde{W}}\right\},
    \label{EQ5.10}
\end{align}
where $\bs{P}$ is a positive definite matrix and $\gamma$ is a positive real number. Then, taking the time derivative of $V$, we obtain 
\begin{align}
\begin{split}
    \largedot{V} &= \Tilde{\bs{z}}^T\left(\bs{A}^T\bs{P} + \bs{PA}\right)\Tilde{\bs{z}} + \bs{\phi}^T\bs{\Tilde{W}}\bs{B}^T\bs{P}\Tilde{\bs{z}} + \Tilde{\bs{z}}^T\bs{P}\bs{B}\bs{\Tilde{W}}^T\bs{\phi}\\
    & \hspace{3ex}+ 2\gamma^{-1}\ts{tr}\left\{\bs{\largedot{\Tilde{W}}}^T\bs{\Tilde{W}}\right\}\\
    &= -\Tilde{\bs{z}}^T\bs{Q}\Tilde{\bs{z}} + 2\ts{tr}\left\{\bs{B}^T\bs{P}\Tilde{\bs{z}}\bs{\phi}^T\bs{\Tilde{W}}\right\}+ 2\gamma^{-1}\ts{tr}\left\{\bs{\largedot{\Tilde{W}}}^T\bs{\Tilde{W}}\right\}\\
    &= -\Tilde{\bs{z}}^T\bs{Q}\Tilde{\bs{z}} + 2\ts{tr}\left\{\bs{B}^T\bs{P}\Tilde{\bs{z}}\bs{\phi}^T\bs{\Tilde{W}}+ \gamma^{-1}\bs{\largedot{\Tilde{W}}}^T\bs{\Tilde{W}}\right\},
    \end{split}
    \label{EQ5.11}
\end{align}
where $\bs{Q}$ is a positive definite matrix that satisfies the Lyapunov equation
\begin{align}
    \bs{A}^T\bs{P} + \bs{PA} = -\bs{Q}.
    \label{EQ5.12}
\end{align}
Accordingly, by choosing the adaptation law
\begin{align}
    \bs{\dot{\hat{W}}}= \bs{\dot{\Tilde{W}}} = -\gamma \bs{\phi} \Tilde{\bs{z}}^T\bs{P}^T\bs{B},
    \label{EQ5.13}
\end{align}
we enforce that
\begin{align}
    \largedot[1.4]{V} = -\Tilde{\bs{z}}^T\bs{Q}\Tilde{\bs{z}} \leq 0,
    \label{EQ5.14}
\end{align}
which implies that the equilibrium \mbox{$\left\{\bs{\Tilde{z}}^{\star}, \bs{\Tilde{W}}^{\star}\right\} = \{\bs{0},\bs{0}\}$} is uniformly stable in the sense of Lyapunov. Therefore, \mbox{$\bs{\Tilde{z}} \in \mathcal{L}_{\infty}$}. 

Next, using \mbox{signal-chasing} analysis, we show the attractivity of $\bs{\Tilde{z}}^{\star}$. First, note that since \mbox{$V(t) \geq 0$} and \mbox{$\largedot{V}(t) \leq 0$},
\begin{align}
    V_{\infty} = \lim_{t\rightarrow \infty}{V(t)} < \infty
    \label{EQ5.15}
\end{align}
and, therefore, this limit exists. Furthermore,
\begin{align}
\begin{split}
    \int_0^{\infty} \largedot{V}(t)dt &= -\int_0^{\infty} \Tilde{\bs{z}}^T\bs{Q}\Tilde{\bs{z}} dt\\
    \Rightarrow V_{\infty} - V(0) &= -\int_0^{\infty} \Tilde{\bs{z}}^T\bs{Q}\Tilde{\bs{z}} dt\\
    \Leftrightarrow V_{\infty} - V(0) &\leq -\lambda_{\ts{min}}\left\{\bs{Q}\right\}\int_0^{\infty} \Tilde{\bs{z}}^T\Tilde{\bs{z}} dt\\
    \Leftrightarrow V_{\infty} - V(0) &\leq -\lambda_{\ts{min}}\left\{\bs{Q}\right\} \|\Tilde{\bs{z}}\|^2_{\mathcal{L}_2}\\
    \Leftrightarrow \|\Tilde{\bs{z}}\|^2_{\mathcal{L}_2} &\leq \frac{V(0) - V_{\infty}}{\lambda_{\ts{min}}\left\{\bs{Q}\right\}} \\
    \Rightarrow \Tilde{\bs{z}} &\in \mathcal{L}_2.
    \end{split}
    \label{EQ5.16}
\end{align}
Since \mbox{$0 \leq V(t)\leq V(0)$} and \mbox{$\bs{\Tilde{z}} \in \mathcal{L}_{\infty}$}, from~\eqref{EQ5.10}, it follows that \mbox{$\bs{\Tilde{W}}(t) \in \mathcal{L}_{\infty}$}, which implies that \mbox{$\bs{\hat{W}}(t) \in \mathcal{L}_{\infty}$}. 
After having shown that all the signals on the right in~\eqref{EQ5.09} are bounded, we conclude that \mbox{$\bs{\dot{\Tilde{z}}} \in \mathcal{L}_{\infty}$}. Thus, all the hypotheses of \mbox{Lemma\,3.2.5} in~\cite{Ioannou_AdaptiveControl} are satisfied and, therefore, it follows that
\begin{align}
    \Tilde{\bs{z}} \rightarrow \bs{0} \quad \ts{as} \quad t \rightarrow \infty.
    \label{EQ5.19}
\end{align}
As a consequence, from~\eqref{EQ5.13}, it also follows that
\begin{align}
    \bs{\dot{\hat{W}}} \rightarrow \bs{0}_{n\times 3} \quad \ts{as} \quad t \rightarrow \infty.
    \label{EQ5.18}
\end{align}
In summary, all the signals of the \mbox{closed-loop} system remain bounded during operation. Additionally, as intended, $\bs{z}$ converges to $\bs{z}_{\ts{r}}$ asymptotically. Note that the convergence of the adaptive matrix $\bs{\hat{W}}$ to $\bs{W}$ is not guaranteed; however, this is not a necessary condition for stability nor for performance improvement as empirically shown in Section\,\ref{Section04}. The block diagram of the MRAC control scheme is shown in Fig.\,\ref{Fig02}.
\begin{figure}[t!]
    \centering
    \includegraphics[width=3.4in]{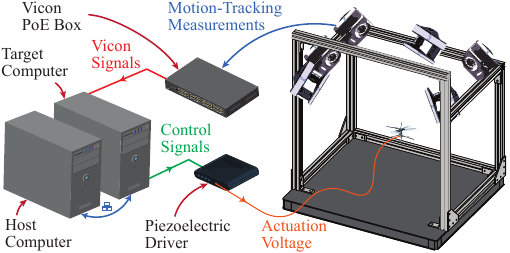}
    \caption{\textbf{Experimental setup used in the \mbox{real-time} flight experiments.} In this
setup, the six DOFs of the robot are measured using a Vicon \mbox{motion-capture} system. The sensor data (position and attitude) are transmitted to a real-time digital controller at a rate of $500$\,Hz. The control algorithms are run in real time on a host-target system, using MathWorks SRT
software, at a fixed time step of $0.5$\,ms. The control signals are mapped into actuator voltages using specialized piezoelectric drivers.}
    \label{Fig03}
\end{figure}
\begin{figure*}[t!]
    \centering
    \includegraphics[width=\textwidth]{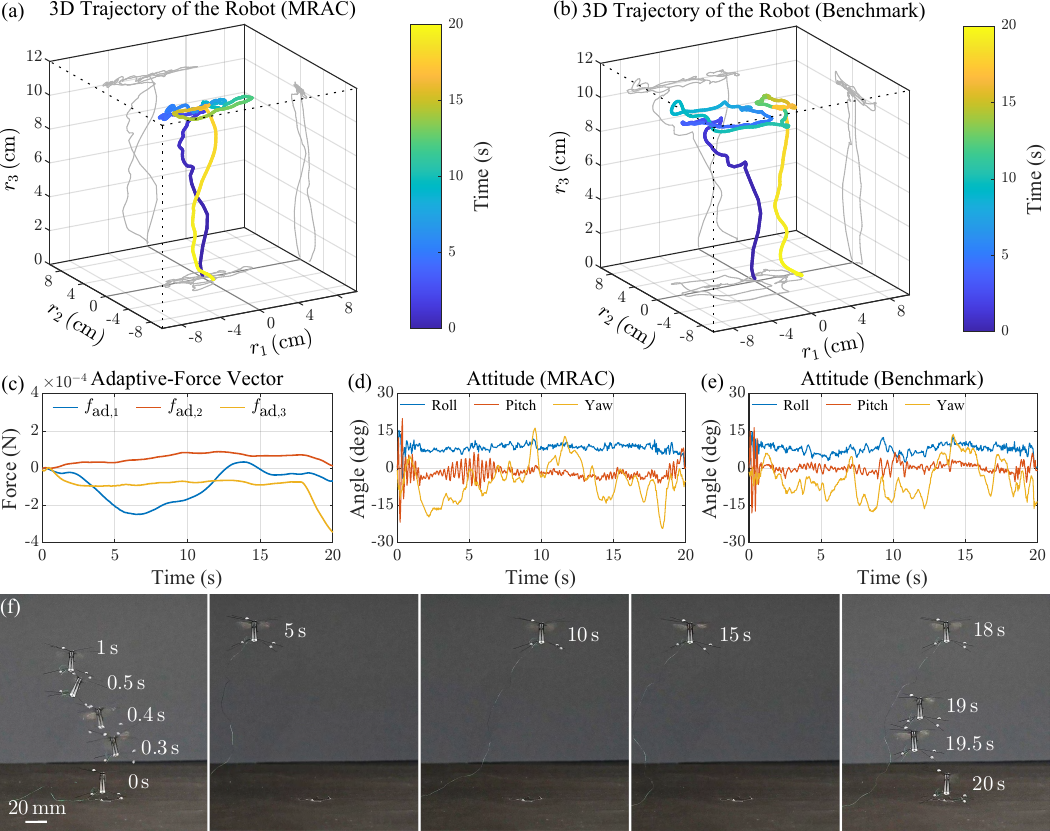}
    \caption{\textbf{Hovering-flight experimental results with yaw regulation.} \textbf{(a)} $3$D trajectory of a \mbox{$20$-second} hovering experiment with a constant reference trajectory \mbox{$\bs{r}_{\ts{d}} = [0~0~10]^T$\,cm}, using the MRAC scheme. \textbf{(b)} $3$D trajectory of a \mbox{$20$-second} hovering experiment with a constant reference trajectory \mbox{$\bs{r}_{\ts{d}} = [0~0~10]^T$\,cm}, using the benchmark control scheme. \textbf{(c)} Time evolution of the three components of the adaptive-force vector, $\bs{f}_{\hspace{-0.2ex}\ts{ad}}$, corresponding to the experiment shown in (a). \textbf{(d)} Time evolution of the three attitude DOFs corresponding to the experiment shown in (a). \textbf{(e)} Time evolution of the three attitude DOFs corresponding to the experiment shown in (b). \textbf{(f)} Photographic composite of frames of the experiment corresponding to the data in (a). Video footage of these experiments is available in the supplementary movie available at \url{https://wsuamsl.com/resources/MovieMRAC.mp4}.}
    \label{Fig04}
    \vspace{-1ex}
\end{figure*}

\section{Experimental Results}\label{Section04}

\subsection{Experimental Setup}\label{Section03-A}
The flying arena used in the flight tests discussed in Section\,\ref{Section03-C} and \ref{Section03-D} is depicted in~Fig.\,\ref{Fig03}. This setup is instrumented with a \mbox{six-V5-camera} Vicon \mbox{motion-capture} system. These cameras measure the position and attitude of the robot during flight, and their data are then streamed, employing \mbox{Tracker\,9.0}, to the \mbox{host--target} system at a rate of \mbox{$500$\,Hz} for monitoring, feedback control, and collection. Before performing a flight experiment, we install retroreflective markers on the tested prototype for motion tracking and place the robot in the arena.  The control and data-processing algorithms are implemented and run in real time on the target computer, using MathWorks SRT software, at a fixed rate of $2$\,kHz. The controller outputs, in the form of actuator signals---mapped as described in~\cite{BenaRM2023I}---are then analogously transmitted from the target computer to the signal generator. This generator, using a piezoelectric driver, amplifies the actuator voltages and provides the power required to properly drive the robot during controlled flight. 

\subsection{Attitude State Estimation}\label{Section03-B}
The Vicon \mbox{motion-capture} system used for sensing can only provide the direct measurements of the robot's attitude, but not its attitude rate. Previously~\cite{BenaRM2022}, researchers have estimated the robot's current and desired attitude rates using a \mbox{low-pass} derivative filter with the form 
\begin{align}
    D(s) = \frac{\lambda s}{s+\lambda},
\end{align}
where $\lambda$ determines the speed of the filter and $s$ is the Laplace complex variable. However, this approach results in extremely noisy estimates, especially of the desired attitude rate, since the corresponding desired attitude is computed by the position controller, as explained in~\cite{BenaRM2022, BenaRM2023I}. Thus, in the experiments reported in this paper, we combine the theory in~\cite{markley2014fundamentals} and~\cite{SimonStateEstimation} to implement a hybrid multiplicative extended K\'{a}lm\'{a}n filter to estimate both the current and desired attitude rates of the robot. In this approach, we model the dynamics of the system in continuous time and its measurements in discrete time. Namely,
\begin{align}
\begin{split}
    \bs{\dot{\zeta}}(t) &= \bs{g}(\bs{\zeta}, \bs{\tau}) + \bs{w}_1(t),\\
    \bs{y}(k) &= \bs{H}\bs{\zeta}(k) + \bs{w}_2(k),
    \end{split}
\end{align}
with \mbox{$\bs{\zeta} = \left[\bs{\qbar}^T~\bs{\omega}^T\right]^T$}, where $\bs{\qbar}$ is a unit quaternion that stores the attitude of the robot and $\bs{\omega}$ is the angular velocity; $\bs{g}(\cdot)$ is a vector-valued function for the attitude dynamics without process noise; $\bs{w}_1$ is \mbox{zero-mean} process noise assumed to have Gaussian distribution with covariance $\bs{R}_1$; \mbox{$\bs{H} = \left[\bs{I}_{4\times 4}~\bs{0}_{4\times 3}\right]$}; and $\bs{w}_2$ is \mbox{zero-mean} measurement noise assumed to have Gaussian distribution with covariance $\bs{R}_2$. In practice, we initialize the filter at $k=0$ with the attitude state estimate and estimation error covariance matrix,
\begin{align}
    \begin{split}
        \bs{\hat{\zeta}}^{+}_0 &= E\left[\bs{\zeta}_0\right],\\
        \bs{C}_0^{+} &= E\left[\left(\bs{\zeta}_0-\bs{\hat{\zeta}}_0\right)\left(\bs{\zeta}_0-\bs{\hat{\zeta}}_0\right)^T\right],
    \end{split}
\end{align}
where $E[\cdot]$ computes the expected value of a random variable. Then, for $k > 0$, we predict the estimate of the state and the estimation error covariance matrix from $(k-1)^+$ to $k^-$ using a noiseless version of the model. Specifically,
\begin{align}
\begin{split}
    \bs{\dot{\hat{\zeta}}} &= \bs{g}(\bs{\hat{\zeta}}, \bs{\tau}),\\
    \bs{\dot{C}} &= \bs{G}\bs{C} + \bs{C}\bs{G}^T + \bs{R}_1,
    \end{split}
\end{align}
where $\bs{G} = \left.\frac{\partial \bs{g}}{\partial \bs{\zeta}}\right|_{\bs{\hat{\zeta}}}$. As explained in~\cite{SimonStateEstimation}, this prediction process is initiated with $\bs{\hat{\zeta}} = \bs{\hat{\zeta}}^{+}_{k-1}$ and $\bs{C} = \bs{C}_{k-1}^{+}$ and at the end of the prediction, we get $\bs{\hat{\zeta}} = \bs{\hat{\zeta}}^{-}_{k}$ and $\bs{C} = \bs{C}_{k}^{-}$. Last, at time index $k$, we use the measurement $\bs{y}_k$ to update the K\'{a}lm\'{a}n gain, the state estimate, and the estimation error covariance matrix according to
\begin{align}
\begin{split}
    \bs{K}_k &= \bs{C}_{k}^{-} \bs{H}^T\left(\bs{H}\bs{C}_{k}^{-}\bs{H}^T + \bs{R}_{2, k}\right)^{-1},\\
    \bs{\hat{\zeta}}_{k}^{+} &= \bs{\hat{\zeta}}_{k}^{-} + \bs{K}_k\left(\bs{y}_k - \bs{H}\bs{\hat{\zeta}}_{k}^{-}\right),\\
    \bs{C}_k^{+} &= (\bs{I} - \bs{K}_{k}\bs{H})\bs{C}_k^{-}.
    \end{split}
\end{align}
For estimating the desired attitude rate, we employ the same method, except that the measured state is the desired attitude computed by the position controller as explained in~\cite{BenaRM2023I}.

\begin{figure}[t!]
    \centering
    \includegraphics[width=3.4in]{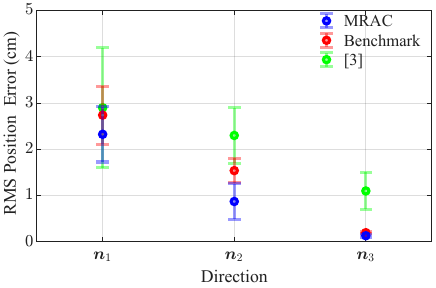}
    \caption{\textbf{Hovering-flight comparison.} Comparison of the mean and ESD of the RMS position error of the robot along the $\bs{n}_1$, $\bs{n}_2$, and $\bs{n}_3$ directions corresponding to the MRAC (blue) and benchmark (red) control approaches using the same prototype, and benchmark approach reported in previously published results (green)~\cite{BenaRM2023I}.}
    \label{Fig05}
\end{figure}
\subsection{Hovering-Flight Experiments}\label{Section03-C}
In the experiments presented in this section, we selected the controller gains in~\eqref{EQ5.04} by using an LQR design approach followed by experimental tuning. Specifically, we used $\bs{K}_{\ts{p}} = \ts{diag}\{1.8,1.8,2\}\times 10^{-2}\,\ts{N}\cdot\ts{m}^{-1}$; $\bs{K}_{\ts{i}} = \ts{diag}\{5,5,15\}\hspace{-0.2ex}\times\hspace{-0.2ex} 10^{-3}\,\ts{N}\hspace{-0.2ex}\cdot\hspace{-0.2ex}\ts{m}^{-1}\hspace{-0.2ex}\cdot\hspace{-0.2ex}\ts{s}^{-1}$; and $\bs{K}_{\ts{d}} = \ts{diag}\{2,2,5\}\hspace{-0.2ex}\times\hspace{-0.2ex} 10^{-3}\,\ts{N}\hspace{-0.2ex}\cdot\hspace{-0.2ex}\ts{m}^{-1}\hspace{-0.2ex}\cdot\hspace{-0.2ex}\ts{s}$. For the~adaptation law in~\eqref{EQ5.13}, we selected $\gamma = 5\cdot10^{-4}$; $\bs{\phi}$ to be a vector with $n=40$ entries, $\bs{c}_i\in [-5,5]\cdot \bs{1}_{n\times 1}$, and $\sigma_i \in [0.01,5]$; and $\bs{P}$ to be the solution to~\eqref{EQ5.12} with
\begin{align}
    \bs{Q} = \begin{bmatrix}
    \bs{0}_{3\times 3} & \bs{0}_{3\times 3} & \bs{0}_{3\times 3}  \\
    \bs{0}_{3\times 3} &\ts{diag}\{1,1,50\}\cdot 10^{-2} & \bs{0}_{3\times 3}\\
    \bs{0}_{3\times 3}& \bs{0}_{3\times 3} & \ts{diag}\{1,1,50\}\cdot 10^{-4}
    \end{bmatrix}.
\end{align}
The first diagonal block of $\bs{Q}$ is set to zero because the additional state \mbox{$\bs{\xi} = \int_0^t \bs{r}_{\ts{e}}(\zeta)d\zeta$} is assumed to have negligible influence on the disturbance model.

To demonstrate the improved performance of the MRAC approach relative to its nonadaptive version described in~\cite{BenaRM2023I}, we executed five \mbox{back-to-back} hovering experiments with each controller, where the desired position was set to \mbox{$\bs{r}_{\ts{d}} = [0~0~10]^T$\,cm}. As shown in Fig.\,\ref{Fig04}, both controllers were able to robustly stabilize the position of the robot; however, it is clear that the experiment where the MRAC approach was implemented (Fig.\,\ref{Fig04}(a)) resulted in a significant better performance compared to the experiment where the benchmark controller was implemented (Fig.\,\ref{Fig04}(b)). Video footage of these experiments is available in the supplementary movie available at \url{https://wsuamsl.com/resources/MovieMRAC.mp4}.
\begin{figure*}[t!]
    \centering
    \includegraphics[width=\textwidth]{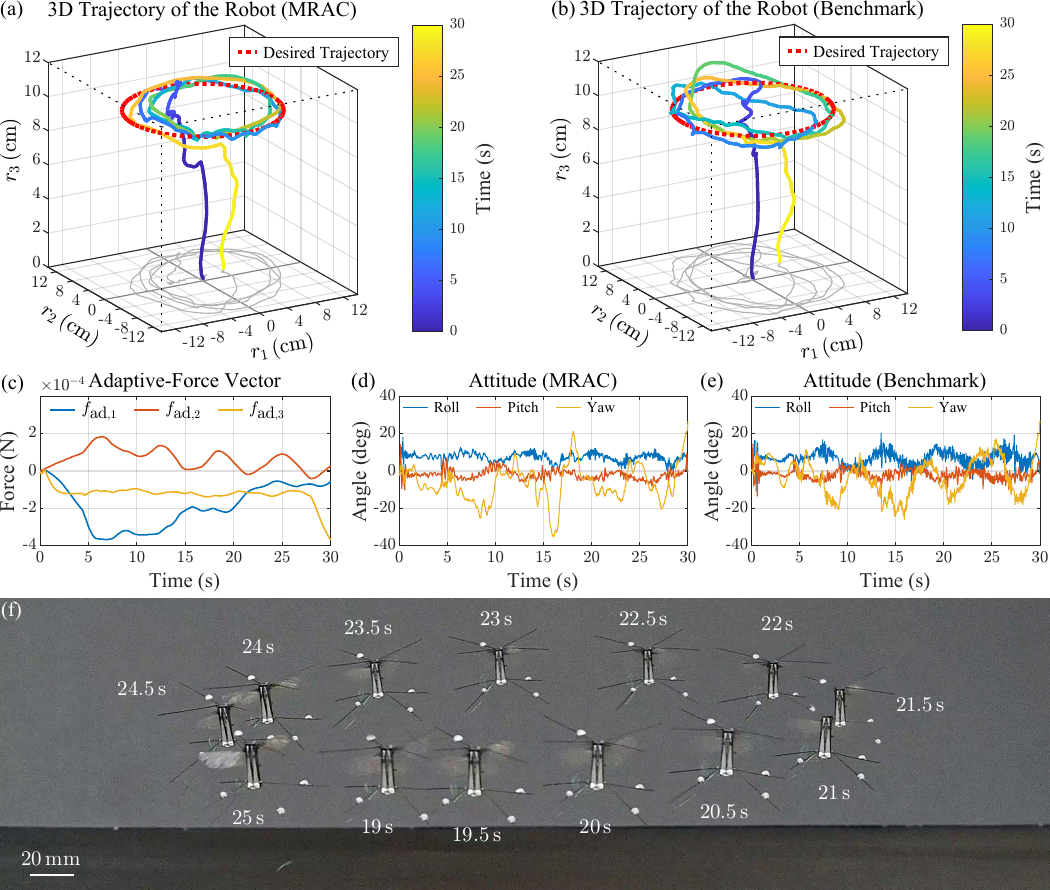}
    \caption{\textbf{Trajectory-tracking experimental results with yaw regulation.} \textbf{(a)} $3$D trajectory of a \mbox{$30$-second} experiment with a time-varying reference trajectory \mbox{$\bs{r}_{\ts{d}} = [10\cos(t-t_0)~10\sin(t-t_0)~10]^T$\,cm} using the MRAC scheme. \textbf{(b)} $3$D trajectory of a \mbox{$30$-second} with the same time-varying reference trajectory as in (a), using the benchmark control scheme. \textbf{(c)} Time evolution of the three components of the adaptive-force vector, $\bs{f}_{\hspace{-0.2ex}\ts{ad}}$, corresponding to the experiment shown in (a). \textbf{(d)} Time evolution of the three attitude DOFs corresponding to the experiment shown in (a). \textbf{(e)} Time evolution of the three attitude DOFs corresponding to the experiment shown in (b). \textbf{(f)} Photographic composite of frames of the experiment corresponding to the data in (a). Video footage of these is available in the supplementary movie available at \url{https://wsuamsl.com/resources/MovieMRAC.mp4}. \label{Fig06}}
\end{figure*}
As seen by the time evolution of the \mbox{adaptive-force} vector shown in Fig.\,\ref{Fig04}(c), the MRAC approach was able to effectively compensate for disturbances and significantly reduce the position-control error. Specifically, the \textit{mean $\pm$ experimental standard deviation}~(ESD) values for the RMS position error corresponding to the experiments where we implemented the MRAC scheme were $2.32\pm0.60$\,cm along the $\bs{n}_1$ direction, $0.87\pm0.39$\,cm along the $\bs{n}_2$ direction, and $0.13\pm0.03$\,cm along the $\bs{n}_3$ direction. For comparison, the values corresponding to the experiments where we implemented the benchmark scheme were $2.74\pm0.63$\,cm along the $\bs{n}_1$ direction, $1.54\pm0.26$\,cm along the $\bs{n}_2$ direction, and $0.20\pm0.02$\,cm along the $\bs{n}_3$ direction. 

\mbox{Figs.\,\ref{Fig04}(d)~and~(e)} show the time evolutions of the three attitude DOFs corresponding to the experiments implementing the MRAC and the benchmark schemes, respectively. Here, it can be observed that all attitude DOFs are effectively stabilized during flight; the bias observed in the roll DOF is most likely caused by a constant offset torque due to small misalignments in fabrication or small errors in the definition of the \mbox{body-fixed} frame in Tracker\,9.0. Furthermore, compared to the results and data reported in~\cite{BenaRM2023I}, the vibrations along these three DOFs are significantly reduced with the implementation of the extended K\'{a}lm\'{a}n filter described in~Section\,\ref{Section03-B}. To empirically verify this claim, we first define $\nu$ as a performance vibration metric. This metric is computed by first calculating the standard deviation values of the roll, pitch, and yaw DOFs for each of the five experiments and then computing the mean of these five ESD values. Specifically,
\begin{align}
    \nu = \frac{\sum^{5}_{i=1}\sigma_i}{5},
\end{align}
where
\begin{align}
    \sigma_i = \sqrt{\frac{\sum^{N}_{j=1}\left(a-\bar{a}\right)^2}{N-1}},
\end{align}
with $N$ being the number of data points in experiment $i$; \mbox{$a\in\{\phi, \theta,\psi\}$} being one of the three attitude DOFs; and, $\bar{a}$ being the mean value of $a$ in experiment $i$. For the experiments reported in~\cite{BenaRM2023I}, the values of $\nu$ for roll, pitch, and yaw were $5.77$\textdegree, $3.82$\textdegree, and $8.77$\textdegree, respectively. For the experiments reported in this paper, the values of $\nu$ for roll, pitch, and yaw were $1.46$\textdegree, $1.69$\textdegree, and $6.80$\textdegree, respectively; in summary, this represents a $75$\%, $56$\%, and $22$\% reduction in vibration along the respective DOFs. As can be observed in the supplementary movie, this fact is also noticeable when comparing the video footage of the \mbox{hovering-flight} experiments in this paper with that of the \mbox{hovering-flight} experiments in~\cite{BenaRM2023I}. Fig.\,\ref{Fig04}(f) shows a photographic composite of frames of the experiment corresponding to the data in Fig.\,\ref{Fig04}(a).

\mbox{Fig.\,\ref{Fig05}} summarizes the results of the \mbox{hovering-flight} experiments by showing a comparison of the mean (solid circle) and ESD (vertical bar) of the RMS position error of the robot corresponding to the MRAC (blue) and benchmark (red) control approaches. Overall, relative to the implementation of the benchmark controller in this prototype, the data corresponding to the MRAC approach represent a mean RMS position error reduction of $15$\,\%, $44$\,\%, and $35$\,\% in the $\bs{n}_1$, $\bs{n}_2$, and $\bs{n}_3$ directions, respectively. Additionally, compared to the previous results reported in~\cite{BenaRM2023I}, the data corresponding to the MRAC approach represent a mean RMS position error reduction of $20$\,\%, $62$\,\%, and $88$\,\% in the $\bs{n}_1$, $\bs{n}_2$, and $\bs{n}_3$ directions, respectively.

\subsection{Trajectory-Tracking Flight Experiments}\label{Section03-D}
To further demonstrate the suitability of the proposed MRAC approach, we conducted additional \mbox{real-time} experiments where the \mbox{insect-scale} prototype was commanded to track a time-varying desired trajectory using the same controller gains as the ones specified in Section\,\ref{Section03-C}. An experiment of this type consists of a take-off maneuver, tracking a circle with a radius of $10$\,cm whose reference position trajectory is defined as \mbox{$\bs{r}_{\ts{d}} = [10\cos(t-t_0)~10\sin(t-t_0)~10]^T$\,cm}---where $t_0$ is the initial time of the circle trajectory---and a landing maneuver. In this case, we executed three \mbox{back-to-back} experiments with each controller, the MRAC and its nonadaptive version used as a benchmark. As shown in Fig.\,\ref{Fig06}, both controllers were able to robustly track the desired trajectory; however, it is clear that the experiment where the MRAC approach was implemented (Fig\,\ref{Fig06}(a)) resulted in a significantly better performance compared to the experiment where the benchmark controller was implemented (Fig\,\ref{Fig06}(b)), where the robot consistently trimmed the desired circular trajectory (dotted red line). Video footage of these experiments is available in the supplementary movie available at \url{https://wsuamsl.com/resources/MovieMRAC.mp4}.

Once again, as seen by the time evolution of the \mbox{adaptive-force} vector shown in Fig.\,\ref{Fig06}(c), the MRAC approach was able to effectively compensate for disturbances and significantly reduce the position-control error. Specifically, the \textit{mean $\pm$ ESD} values for the RMS error corresponding to the experiments where we implemented the MRAC scheme were $3.36\pm0.07$\,cm along the $\bs{n}_1$~direction, $1.32\pm0.17$\,cm along the $\bs{n}_2$~direction, and $0.25\pm0.04$\,cm along the $\bs{n}_3$~direction. For comparison, the values corresponding to the experiments where we implemented the benchmark scheme were $3.54\pm0.26$\,cm along the $\bs{n}_1$~direction, $2.09\pm0.19$\,cm along the $\bs{n}_2$~direction, and $0.35\pm0.07$\,cm along the $\bs{n}_3$~direction. \mbox{Figs.\,\ref{Fig06}(d)~and~(e)} show the time evolutions of the three attitude DOFs corresponding to the experiments implementing the MRAC and the benchmark schemes, respectively. Here, it can be observed that all attitude DOFs are effectively stabilized during flight; as in the case of the \mbox{hovering-flight} experiments, the bias observed in the roll DOF is most likely caused by a constant offset torque due to small misalignments in fabrication or small errors in the definition of the \mbox{body-fixed} frame in Tracker\,9.0. Fig.\,\ref{Fig06}(f) shows a photographic composite of frames of the experiment corresponding to the data in Fig.\,\ref{Fig06}(a).

\begin{figure}[t!]
    \centering
    \includegraphics[width=3.4in]{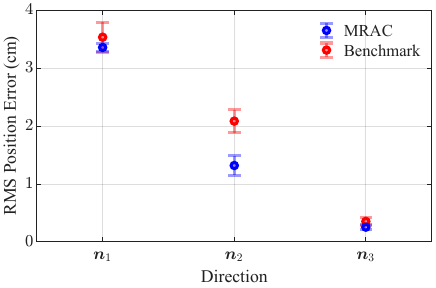}
    \caption[\textbf{Trajectory-tracking performance comparison.}]{\textbf{Trajectory-tracking performance comparison.} Comparison of the mean and ESD of the RMS position error of the robot corresponding to the MRAC (blue) and benchmark (red) control approaches.}
    \label{Fig07}
\end{figure}

\mbox{Fig.\,\ref{Fig07}} summarizes the results of the \mbox{trajectory-tracking} experiments by showing a comparison of the mean and ESD of the RMS position error of the robot corresponding to the MRAC (blue) and benchmark (red) control approaches. In this case, relative to the implementation of the benchmark controller, the data corresponding to the MRAC approach represent a mean RMS position error reduction of $5$\,\%, $37$\,\%, and $28$\,\% in the $\bs{n}_1$, $\bs{n}_2$, and $\bs{n}_3$ directions, respectively. 

We speculate that the larger RMS position error values along the $\bs{n}_1$~direction---relative to the ones in the $\bs{n}_2$~direction---consistently observed in \mbox{Figs.\,\ref{Fig05}~and~\ref{Fig07}} can be explained by the robot's wing arrangement. Specifically, the four wings of the robot are configured in pairs; when actuated at a frequency of $150$\,Hz, this results in the generation of the \mbox{\textit{clap-and-fling}} mechanism along the $\bs{b}_2$~direction, but not along the $\bs{b}_1$~direction as can be observed in the \mbox{high-speed} footage shown in the accompanying supplementary movie. This mechanism has been shown to increase the average thrust by as much as $16$\%~\cite{PhanHV2016}, which therefore suggests that this robot can generate significantly higher forces---and consequently, higher torques---in the $\bs{b}_2$~direction compared to the $\bs{b}_1$~direction giving the system more control authority in that direction.

\section{Conclusion}\label{Section05}
We presented the design and implementation of an MRAC architecture for \mbox{insect-scale} \mbox{flapping-wing} aerial robots. Additionally, we demonstrated that the use of an Extended K\'{a}lm\'{a}n filter to estimate the current and desired angular rates significantly dampens roll, pitch, and yaw vibrations and improves performance. To validate the functionality and effectiveness of the approach, we conducted hovering and \mbox{trajectory-tracking} $6$-DOF flight control experiments with a \mbox{$95$-mg} aerial robot using the MRAC approach and its nonadaptive version, used as a benchmark. The experimental results demonstrate that the MRAC approach yields a significant performance improvement, as measured by the RMS position error. Consistently, this was also the case when the new data were compared to the values reported in~\cite{BenaRM2023I}. For future work, we want (i)\,to extend this approach to the attitude dynamics of the robot, (ii)\,take advantage of the unique capabilities of this robot to control its yaw DOF and implement a collision-avoidance system based on information captured by a virtual camera aligned with its heading, and (iii)\,improve the robot's design to achieve the \mbox{clap-and-fling} mechanism in the $\bs{b}_1$ and $\bs{b}_2$ directions.

\balance
\bibliographystyle{IEEEtran}
\bibliography{references}

\end{document}